\documentclass{article}
\usepackage{ijcai22}

\usepackage{xcolor}
\usepackage{times}
\usepackage{soul}
\usepackage{url}
\usepackage[hidelinks]{hyperref}
\usepackage[utf8]{inputenc}
\usepackage[small]{caption}
\usepackage{graphicx}
\usepackage{amsmath}
\usepackage{amsthm}
\usepackage{booktabs}
\usepackage{algorithm}
\usepackage{algorithmic}
\newtheorem{theorem}{Theorem}
\newtheorem{corollary}{Corollary}
\newtheorem{lemma}{Lemma}
\newtheorem{definition}{Definition}

\newcommand{\etal}{{et al.~}}       % et al.
\newcommand{\ie}{{i.e.~}}           % i.e.
\newcommand{\wrt}{{w.r.t.~}}         % wrt.,
\newcommand{\ifff}{{iff.~}}         % iff.,
\newcommand{\aka}{{a.k.a.~}}        %a.k.a.

\def\circlerightarrow{\put(2,2.5){\circle{2.5}}\rightarrow}

\def\cirlinecir{\put(0.8,2.8){\circle{2.5}}-\put(-0.9,2.8){\circle{2.5}}}

\def\astrightarrow{\put(0.2,-2.2){*}\rightarrow}
\def\astleftarrow{\leftarrow\put(-5,-2.2){*}}

\makeatletter
\newcommand*{\indep}{%
	\mathbin{%
		\mathpalette{\@indep}{}%
	}%
}
\newcommand*{\nindep}{%
	\mathbin{%                   % The final symbol is a binary math operator
		\mathpalette{\@indep}{\not}% \mathpalette helps for the adaptation
	}%
}
\newcommand*{\@indep}[2]{%
	\sbox0{$#1\perp\m@th$}%        box 0 contains \perp symbol
	\sbox2{$#1=$}%                 box 2 for the height of =
	\sbox4{$#1\vcenter{}$}%        box 4 for the height of the math axis
	\rlap{\copy0}%                 first \perp
	\dimen@=\dimexpr\ht2-\ht4-.2pt\relax
	\kern\dimen@
	{#2}%
	\kern\dimen@
	\copy0 %                       second \perp
}

\title{Ancestral Instrument Method for Causal Inference without Complete Knowledge\thanks{Appendices of the paper are available at \url{https://arxiv.org/abs/2201.03810}.}}
\author{
	Debo Cheng$^1$\and
	Jiuyong Li$^1$\and
	Lin Liu$^1$\and
	Jiji Zhang$^2$\and
	Thuc duy Le$^1$\And
	Jixue Liu$^1$\\
	\affiliations
	$^1$ STEM, University of South Australia, Adelaide, SA, Australia\\
	$^2$ Department of Religion and Philosophy, Hong Kong Baptist University, Hong Kong, China\\
	\emails
	\{debo.cheng,jiuyong.li,liu.liu,thuc.le,jixue.liu\}@unisa.edu.au,
	zhangjiji@hkbu.edu.hk 
}
\begin{document}
	
	\maketitle
	
	\begin{abstract}
		Unobserved confounding is the main obstacle to causal effect estimation from observational data. Instrumental variables (IVs) are widely used for causal effect estimation when there exist latent confounders. With the standard IV method, when a given IV is valid, unbiased estimation can be obtained, but the validity requirement on a standard IV is strict and untestable. Conditional IVs have been proposed to relax the requirement of standard IVs by conditioning on a set of observed variables (known as a conditioning set for a conditional IV). However, the criterion for finding a conditioning set for a conditional IV needs a directed acyclic graph (DAG) representing the causal relationships of both observed and unobserved variables. This makes it challenging to discover a conditioning set directly from data. In this paper, by leveraging maximal ancestral graphs (MAGs) for causal inference with latent variables, we study the graphical properties of ancestral IVs, a type of conditional IVs using MAGs, and develop the theory to support data-driven discovery of the conditioning set for a given ancestral IV in data under the pretreatment variable assumption. Based on the theory, we develop an algorithm for unbiased causal effect estimation with a given ancestral IV and observational data. Extensive experiments on synthetic and real-world datasets demonstrate the performance of the algorithm in comparison with existing IV methods.
	\end{abstract}
	
	\section{Introduction}
	\label{Sec:Intro}
	Inferring the total causal effect of a treatment (\aka exposure, intervention or action) on an outcome of interest is a central problem in scientific discovery and it is essential for decision making in many areas such as epidemiology~\cite{martinussen2019instrumental} and economics~\cite{card1993using,verbeek2008guide,imbens2015causal}. With observational data, a major hurdle to causal effect estimation is the bias caused by confounders. Therefore the unconfoundedness assumption is commonly made by causal inference methods~\cite{imbens2015causal}.

	When there are latent or unobserved confounders, the unconfoundedness assumption becomes unreliable. In this case, the instrumental variable (IV) approach~\cite{card1993using,martens2006instrumental} is considered a powerful way to achieve unbiased causal effect estimation. The IV approach leverages an IV (denoted as $S$), a variable known to be a cause of the treatment $W$, controlling treatment assignment, to deal with unobserved confounding. Given a valid IV, an unbiased estimate of the total causal effect of $W$ on outcome $Y$ can be obtained based on the estimated causal effect of $S$ on $W$ and the estimated causal effect of $S$ on $Y$.
	
	The requirements for a standard IV are very strong and it is impossible to find a standard IV in many applications. For a variable $S$ to be a valid standard IV, it must be a cause of $W$ and satisfy the \emph{exclusion restriction} (\ie the causal effect of $S$ on $Y$ must be only through $W$) and be \emph{exogenous} (\ie $S$ does not share common causes with $Y$)~\cite{martens2006instrumental,imbens2014instrumental}. These conditions are strict and can only be justified by domain knowledge. In particular, the exogeneity implies that $S$ must be a factor ``external'' to the system under consideration and connects to the system only through the treatment $W$, which is impossible to validate in practice.
	
	A conditional IV relaxes the requirements of a standard IV significantly and it is more likely to exist in an application than a standard IV~\cite{pearl2009causality,brito2002generalized}. With the concept of a conditional IV, an ``internal'' variable $S$ can be a valid IV when conditioning on a set of observed variables $\mathbf{Z}$. In this case, $S$ is known as a conditional IV which is instrumentalized by $\mathbf{Z}$, and the key to the success of the conditional IV method (in obtaining unbiased causal effect estimation) is to find a proper conditioning set $\mathbf{Z}$ for a given conditional IV. 
	
	However, the criterion for finding $\mathbf{Z}$ is based on complete causal structure knowledge (i.e. a complete causal DAG with observed and unobserved variables), which, if at all possible, can only be obtained from domain knowledge, not observational data. Moreover, recent work~\cite{van2015efficiently} has shown that the search for $\mathbf{Z}$ in a DAG is NP-hard for a given conditional IV. The authors also proposed the concept of an ancestral IV in a DAG, a restricted version of a conditional IV, to work towards efficient search for  $\mathbf{Z}$. Nonetheless, the search for $\mathbf{Z}$ for an ancestral IV in a DAG still requires a DAG containing all the observed and unobserved variables. Therefore, the majority of existing methods for finding a conditioning set of a conditional IV need a causal graph which may not be known in many applications.

	There are some works which \emph{use} the conditional IV without complete causal knowledge, such as random forest for IV~\cite{athey2019generalized}, an estimator based on the assumption of the existence of some invalid and some valid IVs (sisVIVE)~\cite{kang2016instrumental}, and IV.tetrad~\cite{silva2017learning}, but they do not identify conditioning sets. We differentiate our work from these works in the Related Work section in more detail and compare our method with them in the Experiments section.
	
	In this paper, we design an algorithm for identifying a conditional set that instrumentalizes a given ancestral IV, a type of conditional IVs, in data directly. In order to achieve this, we study the graphical properties of an ancestral IV using a MAG (maximal ancestral graph~\cite{richardson2002ancestral,zhang2008causal}) and develop the theory for data-driven discovery of a conditioning set for a given ancestral IV. To the best of our knowledge, there is no existing method for finding a conditioning set of a conditional IV directly from data.
	
	The contributions of this work are summarized as follows.
	\begin{itemize}
		\item We study the novel graphical properties of an ancestral IV using MAGs, which enables a data-driven approach to applying the IV method to obtain unbiased causal effect estimation when there are latent confounders.
		\item  We establish graphical criteria for determining a conditioning set of a given ancestral IV via a MAG (PAG).
		\item Based on the theorems, we propose an effective algorithm for unbiased causal effect estimation from data with latent variables. The experiments on synthetic and real-world datasets demonstrate the performance of the proposed algorithm.
	\end{itemize}

	\section{Background}
	\label{sec:pre}
	\subsection{Graphical Notation and Definitions}
	\label{subsec:graph}
	A graph $\mathcal{G}\!\!=\!\!(\mathbf{V}, \mathbf{E})$ consists of a set of nodes $\mathbf{V}\!\!=\!\!\{V_{1}, \dots, V_{p}\}$, denoting random variables, and a set of edges $\mathbf{E} \!\subseteq \mathbf{V} \times \mathbf{V}$, representing the relationships between nodes. Two nodes linked by an edge are \textit{adjacent}. In the paper, an edge in $\mathbf{E}$ can be a directed edge $\rightarrow$, a bi-directed edge $\leftrightarrow$, or a partially directed edge $\circlerightarrow$, where the circle at the left end of the edge indicates uncertainty of the orientation. 
	
	A path $\pi$ between $V_{i}$ and $V_{j}$ in a graph comprises a sequence of distinct nodes $\langle V_{i}, \dots, V_{j}\rangle$ with every pair of successive nodes being adjacent. $V_i$ and $V_j$ are end nodes of $\pi$, and other nodes on $\pi$ are non-end nodes. A path is a directed or causal path if all edges along it are directed such as $V_{i} \rightarrow\ldots \rightarrow V_{j}$. We use `$\ast$' to indicate an arbitrary edge mark of an edge, \ie arrow ($>$), tail ($-$) or circle ($\circ$). $V_{i}$ is a collider on a path if $V_{i-1}\astrightarrow V_{i} \astleftarrow V_{i+1}$ is in $\mathcal{G}$. A \emph{collider path} is a path on which every non-endpoint node is a collider. A path of length one is a \emph{trivial collider path}.

	If there is $V_{i}\leftrightarrow V_{j}$ in a graph, $V_i$ and $V_j$ are called spouses to each other. We use $Adj(V)$, $Pa(V)$, $Ch(V)$, $An(V_i)$, $De(V_i)$, $Sp(V)$ and $PossAn(V)$ to denote the sets of all adjacent nodes, parents, children, ancestors, descendants, spouses and possible ancestors of $V$, respectively, in the same way as in~\cite{perkovic2018complete}. The definitions of a node's parents, children, ancestors and descendants are provided in Appendix A~\cite{cheng2022ancestral}. A directed cycle occurs when the first and last nodes on a path are the same node. A \emph{DAG} contains directed edges without cycles. In a DAG with observed and unobserved variables, if there exists $V_{i}\leftarrow U \rightarrow V_{j}$ where $U$ is a latent variable, $V_i$ and $V_j$ are often called spouses to each other.

	Ancestral graphs are often used to represent the mechanisms of the data generation process that may involve latent variables~\cite{zhang2008causal}. An ancestral graph is a graph that does not contain directed cycles or almost directed cycles~\cite{richardson2002ancestral}. An almost directed cycle occurs if $V_i\leftrightarrow V_j$ and $V_j\in An(V_i)$.
	
	To save space, the definitions of Markov property, faithfulness,  d-separation (denoted as  $\indep_{d}$), d-connecting (denoted as  $\nindep_{d}$), m-separation (denoted as $\indep_{m}$), m-connecting (denoted as $\nindep_{m}$), and the graphical criteria of d-separation and m-separation are introduced in Appendix A.
	
	\begin{definition}[MAG]
		\label{MAG}
		An ancestral graph $\mathcal{M}=(\mathbf{V}, \mathbf{E})$ is a MAG when every pair of non-adjacent nodes $V_{i}$ and $V_{j}$ in $\mathcal{M}$ are  m-separated by a set $\mathbf{Z}\subseteq \mathbf{V}\backslash \{V_{i}, V_{j}\}$.
	\end{definition}
	
	A DAG obviously meets the conditions of a MAG, so syntactically, a DAG is also a MAG without bi-directed edges. It is worth noting that a causal DAG over a set of observed and unobserved variables can be converted to a MAG over the observed variables uniquely according to the construction rules~\cite{zhang2008completeness}. A set of Markov equivalent MAGs can be represented uniquely by a \emph{partial ancestral graph} (PAG) that is defined in Appendix A.
	
	\begin{definition}[Visibility~\cite{zhang2008causal}]
		\label{Visibility}
		Given a MAG $\mathcal{M}=(\mathbf{V}, \mathbf{E})$, a directed edge $V_{i}\rightarrow V_{j}$ is visible if there is a node $V_{k}\notin Adj(V_{j})$, such that either there is an edge between $V_{k}$ and $V_{i}$ that is into $V_{i}$, or there is a collider path between $V_{k}$ and $V_{i}$ that is into $V_{i}$ and every node on this path is a parent of $V_{j}$. Otherwise, $V_{i}\rightarrow V_{j}$ is said to be invisible.
	\end{definition}
	
	In a given DAG $\mathcal{G}$, if $V_i$ and $V_j$ are not adjacent and $V_i\notin An(V_j)$, then $Pa(V_i)$ blocks all paths between $V_i$ to $V_j$. In a given MAG $\mathcal{M}$, there is a similar conclusion, but the blocked set is $D$-$SEP(V_i, V_j)$  as defined below, instead of $Pa(V_i)$.
	
	\begin{definition}[$D$-$SEP(V_i, V_j)$ in a MAG $\mathcal{M}$~\cite{spirtes2000causation}]
		\label{def:DSEP}
		In a MAG $\mathcal{M}=(\mathbf{V}, \mathbf{E})$, assume that $V_i$ and $V_j$ are not adjacent. A node $V_k \in D$-$SEP(V_i, V_j)$ if $V_k \neq V_i$, and there is a collider path between $V_k$ to $V_i$ such that every node on this path (including $V_k$) is in $An(V_i)$ or $An(V_j)$ in $\mathcal{M}$.
	\end{definition}
	
	\subsection{Instrumental Variables}
	\label{subsec:IVmethods}
	In this section, we introduce the concepts of standard IVs, conditional IVs and ancestral IVs in a DAG $\mathcal{G}=(\mathbf{V, E})$ with $\mathbf{V}= \mathbf{X}\cup\mathbf{U}\cup\{W, Y\}$, where $\mathbf{X}$ is the set of all observed variables and $\mathbf{U}$ is the set of unobserved variables. 
	\begin{definition}[Standard IV]
		\label{def:conIV}
		A variable $S$ is said to be an IV \wrt $W\rightarrow Y$, if (i) $S$ is a cause of $W$, (ii) $S$ affects $Y$ only through $W$ (\ie exclusion restriction), and (iii) $S$ does not share common causes with $Y$ (\ie $S$ is exogenous).
	\end{definition}
	
\begin{figure}[t]
		\centering
		\includegraphics[scale=0.425]{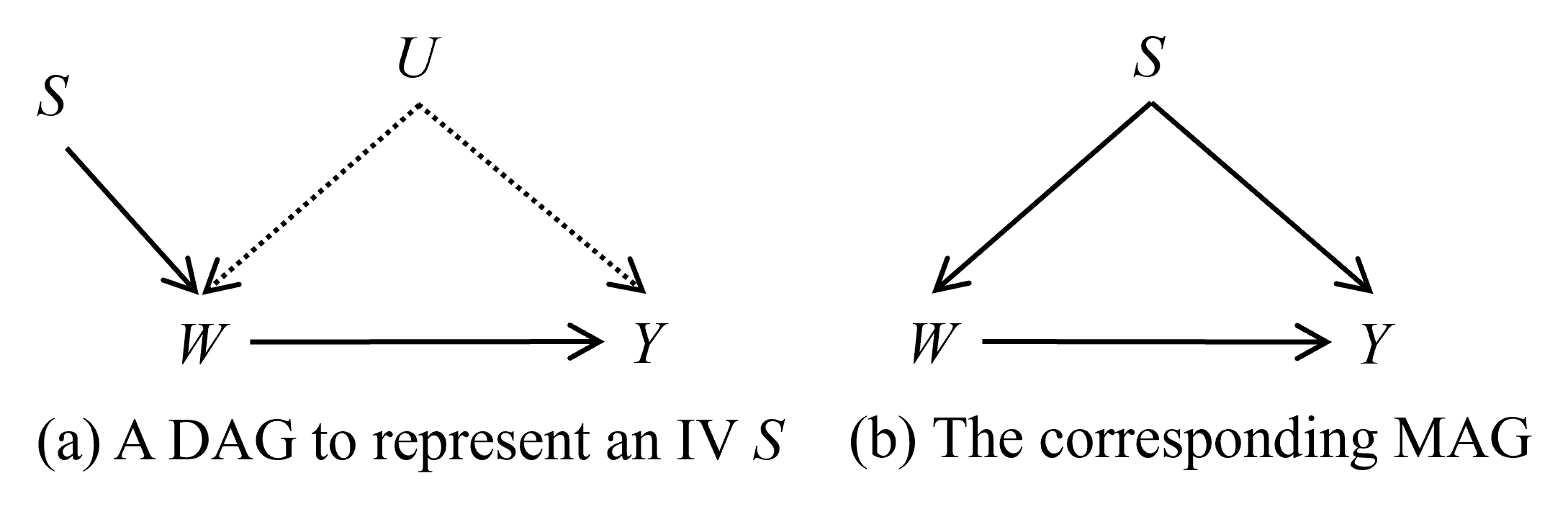}
		\caption{An example of a standard IV represented in two types of causal graphs: (a) DAG; (b) MAG where $W\rightarrow Y$ is invisible.}
		\label{fig:fig001}
	\end{figure}

	The variable $S$ in the DAG in Fig.~\ref{fig:fig001} (a) depicts a standard IV \wrt $W\rightarrow Y$. Given a standard IV $S$, the causal effect of $W$ on $Y$, denoted as $\beta_{wy}$ can be calculated as $\sigma_{sy}/\sigma_{sw}$, where $\sigma_{sy}$ and $\sigma_{sw}$ are the estimated causal effect of $S$ on $Y$ and the causal effect of $S$ on $W$, respectively.

	A conditional IV in a DAG (Definition 7.4.1 on Page 248~\cite{pearl2009causality}) is defined as follows.
	\begin{definition}[Conditional IV]
		\label{def:conditionalIV}
		Given a DAG $\mathcal{G}=(\mathbf{V, E})$ with $\mathbf{V}=\mathbf{X}\cup\mathbf{U}\cup\{W, Y\}$, a variable $S$ is said to be a conditional IV \wrt $W\rightarrow Y$ if there exists a set of observed variables $\mathbf{Z}\subseteq\mathbf{X}$ such that (i) $S\nindep_d W\mid\mathbf{Z}$, (ii) $S\indep_d Y\mid\mathbf{Z}$ in $\mathcal{G}_{\underline{W}}$, and (iii) $\forall Z\in \mathbf{Z}$, $Z\notin De(Y)$.
	\end{definition}
	
	In the above definition, $\mathcal{G}_{\underline{W}}$is the DAG obtained by removing $W\rightarrow Y$ from $\mathcal{G}$. It is worth noting that $\mathbf{Z}$ is a set of observed variables and $\mathbf{Z}\neq \emptyset$ for a conditional IV $S$. 
	
	Detention~\ref{def:conditionalIV} allows a conditional IV $S$ such that $S$ is not related to $W$, but conditioning on $\mathbf{Z}$, is related to $W$ when $\mathbf{Z}$ contains a descendant of $S$. This might lead to a misleading result~\cite{van2015efficiently}. The following defined notion mitigates this issue.
	
	\begin{definition}[Ancestral IV in DAG~\cite{van2015efficiently}]
		\label{def:ancestralIV}
		Given a DAG $\mathcal{G}=(\mathbf{X}\cup \mathbf{U}\cup \{W, Y\}, \mathbf{E})$, a variable $S\in \mathbf{X}$ is said to be an ancestral IV \wrt $W\rightarrow Y$, if there exists a set of observed variables $\mathbf{Z}\subseteq\mathbf{X}\setminus \{S\}$ such that (i) $S\nindep_d W\mid\mathbf{Z}$, (ii) $S\indep_d Y\mid\mathbf{Z}$ in $\mathcal{G}_{\underline{W}}$, and (iii) $\mathbf{Z}\subseteq \{An(Y)\cup An(S)\}$ and $\forall Z\in \mathbf{Z}$, $Z\notin De(Y)$.
	\end{definition}
	
	In a given DAG $\mathcal{G}$, an ancestral IV is a conditional IV, but a conditional IV may not be an ancestral IV. However, the application of a standard IV, conditional IV or ancestral IV requires that a causal DAG $\mathcal{G}=(\mathbf{X}\cup \mathbf{U}\cup \{W, Y\}, \mathbf{E})$ must be completely known. Often, it is impractical to get such complete causal knowledge in real-world applications.

	Using the IV approach, one way to estimate $\beta_{wy}$ from data is to employ the generalized linear model. In this work, we consider the potential outcome model~\cite{imbens2015causal} to calculate $\beta_{wy}$ as introduced in the following. 
	\begin{equation}\label{eq:001}
		\eta\{\mathbf{E}(y\mid w,s,\mathbf{z})\}-\eta\{\mathbf{E}(y_0\mid w,s,\mathbf{z})\} = f^{T}(\mathbf{z})w\beta_{wy}
	\end{equation}
	\noindent where $y, w, s$ and $\mathbf{z}$ denote the values of $Y, W, S$ and $\mathbf{Z}$ respectively for a given individual, $y_0$ is the potential outcome with $w$ set to $0$, and $\eta$ is the identity, log or logit link. The function $f^{T}(\mathbf{z})$ allows us to measure the interactions between $W$ and $\mathbf{Z}$. As commonly done in literature, we utilize a two-stage estimation to estimate  $\beta_{wy}$. The estimator requires two regression models. The first stage is to build a regression model $\hat{w} =\hat{\mathbf{E}}(w\mid s, \mathbf{z})$ for each individual from data. The second stage is to fit the outcome by using $\mathbf{Z}$ and $f^{T}(\mathbf{z})\hat{w}$ as regressors. Hence, the estimated coefficient of $f^{T}(\mathbf{z})\hat{w}$ is $\hat{\beta}_{wy}$. For more details on the estimator, please refer to the literature~\cite{sjolander2019instrumental}.
	
	\section{Finding a Conditioning Set for an Ancestral IV in Data}
	\label{sec:theo}
	\subsection{Problem Setting}
	\label{sec:proset}
	In this work, we assume that an ancestral IV $S$ has been given, and there exists a conditioning set $\mathbf{Z}\subseteq\mathbf{X}\setminus\{S\}$ and $\mathbf{Z}\neq\emptyset$ for $S$ in the underlying DAG $\mathcal{G}$ over $\mathbf{X}\cup \mathbf{U}\cup \{W,Y\}$. We assume that $\mathbf{X}$ contains only pretreatment variables as often assumed in literature~\cite{imbens2015causal,silva2017learning}, \ie for each $X\in\mathbf{X}$, $X \notin De(W)$ and $X\notin De(Y)$ in $\mathcal{G}$. The goal of the work is to provide a practical solution for finding a set of observed variables $\mathbf{Z}$ for a given ancestral IV without knowing the complete causal knowledge. Clause (iii) in Definition~\ref{def:ancestralIV} is too restrictive for finding $\mathbf{Z}$ in data directly because in a PAG, an ancestor and a spouse of a node may not be distinguished. Hence, we consider a relaxed condition for clause (iii) of Definition~\ref{def:ancestralIV}, \ie  $\mathbf{Z}$ does not contain a collider on a d-connecting path between $S$ and $W$ since this is sufficient to address the original problem with the notion of a conditional IV. Hereinafter, we consider that in a DAG $\mathcal{G}$, if a set $\mathbf{Z}\subseteq\mathbf{X}\setminus \{S\}$ satisfies clauses (i) and (ii) in Definition~\ref{def:ancestralIV}, and does not contain a collider between $S$ and $W$, then $\mathbf{Z}$ instrumentalizes $S$.

	Furthermore, we consider the case that an ancestral IV $S$ in a DAG is a cause or spouse of $W$ (\ie a node in $\{Pa(W)\cup Sp(W)\}$) because it is easy to know a cause of $W$ or a spouse of $W$ that is not a direct cause or spouse of $Y$. In the graphic term, $S$ is adjacent to $W$ but not to $Y$. For example, when estimating causal effect of \textit{Smoking} on \textit{Lung Cancer}, \textit{Income} is a direct cause of $Smoking$, but not a direct cause of \textit{Lung Cancer}~\cite{spirtes2000causation}. Hence, \textit{Income} can be used as an IV. It is feasible for users to find an $S$ similar to the case described above. When we infer causal effect from data, we follow the convention in causal inference, that is, the causal DAG $\mathcal{G}$ satisfies Markov property, the causal DAG $\mathcal{G}$ and the data are faithful to each other~\cite{spirtes2000causation,pearl2009causality}. All proofs in this section are provided in Appendix B.

	\subsection{Representing an Ancestral IV in MAG}
	\label{Subsec:relaxedIV}
	An advantage of MAGs is their ability in representing causal relationships between observed variables without involving latent variables that exist in the system~\cite{spirtes2000causation}. A PAG that represents the Markov equivalence class of MAGs can be learned from data with latent variables. The goal of our work is to study the graphical properties of an ancestral IV in a mapped MAG (or equivalently in a PAG) and establish the corresponding theorems for supporting a practical algorithm to estimate $\beta_{wy}$ from data using ancestral IVs.  
	
	When we use a MAG $\mathcal{M}$ over $\mathbf{X}\cup \{W, Y\}$ to represent the data generation mechanism involving latent variables $\mathbf{U}$, an IV in the underlying DAG over $\mathbf{X}\cup\mathbf{U}\cup\{W, Y\}$ can be mapped to $\mathcal{M}$. As all types of IVs (standard, conditional or ancestral IVs) have spurious associations with $Y$ because of the latent confounder between $W$ and $Y$, we develop a lemma for properly mapping an IV in a DAG to a MAG.
	
	\begin{lemma}
		\label{lemma:001}
		Given a DAG $\mathcal{G} = (\mathbf{X}\cup\mathbf{U}\cup\{W, Y\}, \mathbf{E}')$ with the edges $W\rightarrow Y$ and $W\leftarrow U\rightarrow Y$ in $\mathbf{E}'$, and $U\in \mathbf{U}$. Let $\mathcal{M} = (\mathbf{X}\cup\{W, Y\}, \mathbf{E})$ be the MAG mapped from $\mathcal{G}$ based on the construction rules~\cite{zhang2008completeness}. Suppose that there exists an ancestral IV $S$ conditioning on a set  $\mathbf{Z}\subseteq\mathbf{X}\setminus\{S\}$ in $\mathcal{G}$. In the mapped MAG $\mathcal{M}$, the edge $W\rightarrow Y$ is invisible and there is an edge $S\rightarrow Y$ or $S\leftrightarrow Y$. 
	\end{lemma}
	
	We take the standard IV $S$ in the DAG of Fig.~\ref{fig:fig001} (a) as an example to explain the lemma. In this, $S$ is a standard IV \wrt $W\rightarrow Y$ and $S\in An(Y)$, so the IV $S$ in the mapped MAG $\mathcal{M}$ has a directed edge $S\rightarrow Y$ as shown in Fig.~\ref{fig:fig001} (b) and the edge $W\rightarrow Y$ is invisible. 
	
	\subsection{The Property of an Ancestral IV in MAG}
	\label{subsec:property_ANIV}
	First of all, we introduce the manipulated MAG $\mathcal{M}_{\underline{W}\widetilde{S}}$ that is obtained by replacing $W\rightarrow Y$ with $W\leftrightarrow Y$ in $\mathcal{M}$ and removing the edge between $S$ and $Y$. We have the following lemma to present the property of an ancestral IV $S$ in the MAG $\mathcal{M}$ mapped from a DAG $\mathcal{G}$.
	\begin{lemma}[The property of an ancestral IV in the mapped MAG]
		\label{lemma:002}
		Given a DAG $\mathcal{G} = (\mathbf{X}\cup\mathbf{U}\cup\{W, Y\}, \mathbf{E}')$ with the edges $W\rightarrow Y$ and $W\leftarrow U\rightarrow Y$ in $\mathbf{E}'$, and $U\in\mathbf{U}$, and let $\mathcal{M} = (\mathbf{X}\cup\{W, Y\}, \mathbf{E})$ be the MAG mapped from $\mathcal{G}$. Suppose that there exists an ancestral IV $S$ conditioning on a set $\mathbf{Z}\subseteq\mathbf{X}\setminus\{S\}$ in $\mathcal{G}$. In the mapped MAG $\mathcal{M}$, if a set $\mathbf{Z}\subseteq\mathbf{X}\setminus\{S\}$ satisfies the conditions that (i) $S$ and $W$ are not m-separated given $\mathbf{Z}$ in $\mathcal{M}$, and (ii) $S$ and $Y$ are m-separated by $\mathbf{Z}$ in $\mathcal{M}_{\underline{W}\widetilde{S}}$,	then $\mathbf{Z}$ instrumentalizes $S$ in the DAG $\mathcal{G}$.
	\end{lemma}

	\subsection{Determining a Conditioning Set Using a MAG}
	\label{Subsec:relaxedIV}
	The following lemma from the work in~\cite{maathuis2015generalized} is useful for our purpose.
	\begin{lemma}
		\label{lemma:2015}
		Let $X$ and $Y$ be two non-adjacent nodes in a MAG $\mathcal{M}$, then $X\indep_m Y\mid D$-$SEP(X, Y)$.
	\end{lemma}
	
	Therefore, we have the following corollary for finding a set $\mathbf{Z}\subseteq \mathbf{X}\setminus\{S\}$ in a MAG $\mathcal{M}$ that instrumentalizes $S$ in the underlying DAG $\mathcal{G}$.
	
	\begin{corollary}
		\label{theo:IV_MAG}
		Given a DAG $\mathcal{G} = (\mathbf{X}\cup\mathbf{U}\cup\{W, Y\}, \mathbf{E}')$ with the edges $W\rightarrow Y$ and $W\leftarrow U\rightarrow Y$ in $\mathbf{E}'$, and $U\in\mathbf{U}$, and let $\mathcal{M} = (\mathbf{X}\cup\{W, Y\}, \mathbf{E})$ be the MAG mapped from $\mathcal{G}$. For a given ancestral IV $S$, $D$-$SEP(S, Y)$ in the mapped MAG $\mathcal{M}$ is a set that instrumentalizes $S$ in the DAG $\mathcal{G}$.
	\end{corollary}

	Corollary~\ref{theo:IV_MAG} provides a theoretical solution for determining a conditioning set that instrumentalizes a given ancestral IV $S$ in the underlying DAG. Taking a data-driven approach, we can learn a PAG from data with latent variables, but for each of the Markov equivalent MAGs represented by the PAG, there is a corresponding $D$-$SEP(S, Y)$ for $S$. We do not know which MAG is the ground-truth MAG that is mapped from the underlying DAG, and hence we do not know which $D$-$SEP(S, Y)$ is the true conditioning set for $S$. To provide a precise causal effect estimation, in the next section, we propose a theorem to determine a conditioning set $\mathbf{Z}$ from a PAG, in which non-ancestral nodes of $S$ or $Y$ may be contained in $\mathbf{Z}$, but do not result in bias.

	\subsection{Determining a Conditioning Set $\mathbf{Z}$ Using a PAG}
	\label{subsec:conditioningset_PAG}
	For a given ancestral IV $S$, we have the following theorem for determining a set $\mathbf{Z}$ in a PAG $\mathcal{P}$ that instrumentalizes $S$. 
	
	\begin{theorem}
		\label{theo:IV_PAG}
		Given a DAG $\mathcal{G} = (\mathbf{X}\cup\mathbf{U}\cup\{W, Y\}, \mathbf{E}')$ with the edges $W\rightarrow Y$ and $W\leftarrow U\rightarrow Y$ in $\mathbf{E}'$, and $U\in\mathbf{U}$, and let $\mathcal{M} = (\mathbf{X}\cup\{W, Y\}, \mathbf{E})$ be the MAG mapped from $\mathcal{G}$. From data, the mapped MAG $\mathcal{M}$ is represented by a PAG $\mathcal{P}=(\mathbf{X}\cup\{W, Y\}, \mathbf{E}'')$. For a given ancestral IV $S$ which is a cause or spouse of $W$, the set $PossAn(S\cup Y)\setminus\{W, S\}$ in the learned $\mathcal{P}$ is a set that instumentalizes $S$ in the DAG $\mathcal{G}$.
	\end{theorem}
	
	\noindent Note that $PossAn(S\cup Y)\setminus\{W, S\}$ is a superset of $D$-$SEP(S, Y)$ since the former may contain non-ancestral nodes of $S$ or $Y$, and do not result in bias. Theorem~\ref{theo:IV_PAG} allows us to discover the conditioning set as $PossAn(S\cup Y)\setminus \{W, S\}$ from the manipulated PAG $\mathcal{P}_{\underline{W}\widetilde{S}}$ for the given ancestral IV $S$ without complete causal knowledge.

	\subsection{The Proposed Ancestral IV Based Estimator}
	\label{subsec:datadrimeth}
	In this section, we develop a data-driven estimator, \underline{A}ncestral \underline{IV} estimator \underline{i}n \underline{P}AG (\emph{AIViP}) as shown in Algorithm~\ref{pseudocode01}, for unbiased causal effect estimation with a given ancestral IV and data with latent confounders. 
	
	\begin{algorithm}[t]
		\caption{\underline{A}ncestral \underline{IV} estimator \underline{i}n \underline{P}AG (\emph{AIViP})}
		\label{pseudocode01}
		\noindent {\textbf{Input}}: Dataset $\mathbf{D}$ with the treatment $W$, the outcome $Y$, the set of pretreatment variables $\mathbf{X}$ and ancestral IV $S$\\
		\noindent {\textbf{Output}}: $\hat{\beta}_{wy}$ 
		\begin{algorithmic}[1]
			\STATE {Call the causal structure learning method, rFCI, to learn a PAG $\mathcal{P}$ from $\mathbf{D}$}
			\STATE{Obtain the manipulated PAG $\mathcal{P}_{\underline{W}\widetilde{S}}$}
			\STATE{Obtain the set $PossAn(S\cup Y)\setminus \{W, S\}$ in $\mathcal{P}_{\underline{W}\widetilde{S}}$}
			\STATE{$\mathbf{Z} = PossAn(S\cup Y)\setminus \{W, S\}$} 
			\STATE {fit $\hat{w} =\hat{\mathbf{E}}(w\mid s, \mathbf{z})$}
			\STATE {fit $\hat{y}= \hat{\mathbf{E}}(y\mid f^{T}(\mathbf{z})\hat{w}, \mathbf{z})$}
			\STATE {Let $\hat{\beta}_{wy}$ be the coefficient of $f^{T}(\mathbf{z})\hat{w}$}
			\RETURN{$\hat{\beta}_{wy}$}
		\end{algorithmic}
	\end{algorithm}
	
	As presented in Algorithm~\ref{pseudocode01}, \emph{AIViP} (Line 1) employs a global causal structure learning method to discover a PAG from data with latent variables. In this work, we employ rFCI~\cite{colombo2012learning} and use the function \emph{rfci} in the $\mathbf{R}$ package \emph{pcalg}~\cite{kalisch2012causal} to implement rFCI. Lines 2 and 3 construct the manipulated PAG $\mathcal{P}_{\underline{W}\widetilde{S}}$ and get $PossAn(S\cup Y)\setminus \{W, S\}$ from the PAG $\mathcal{P}_{\underline{W}\widetilde{S}}$. Lines 4 to 7 estimate $\hat{\beta}_{wy}$ by the two-stage regression in Eq.(\ref{eq:001}). We use the function \emph{glm} in the $\mathbf{R}$ package \emph{stats} to fit $\hat{w}$ and the function \emph{ivglm} in the $\mathbf{R}$ package \emph{ivtools}~\cite{sjolander2019instrumental} to fit $\hat{y}$.

	\section{Experiments}
	\label{sec:Exp}
	The goal of the experiments is to evaluate the performance of \emph{AIViP} in obtaining the causal effect estimate $\hat{\beta}_{wy}$, especially, when there is a latent confounder between $W$ and $Y$. Five benchmark causal effect estimators are used in the comparison experiments, including  the IV.tetrad in~\cite{silva2017learning}; some invalid some valid IV estimator (sisVIVE)~\cite{kang2016instrumental}; two-stage least squares for standard IV (TSLS)~\cite{angrist1995two}, the most popular estimator; An extension of TSLS for a conditional IV by conditioning on the set of all variables $\mathbf{X}\setminus\{S\}$ (TSLSCIV)~\cite{imbens2014instrumental}; the causal random forest for IV regression (FIVR), with a given conditional IV~\cite{athey2019generalized}.
	
	It is worth noting that since IV.tetrad is the only other data-driven conditional IV method, we also include the data-driven standard IV method (sisVIVE), the most popular standard IV method (TSLS) and its extension to condition IVs (TSLSCIV), and the popular random forest based estimator with a given conditional IV (FIVR).
	
	\paragraph{Implementation and Parameter Setting.} The implementation of IV.tetrad is retrieved from the authors' site\footnote{\url{ http://www.homepages.ucl.ac.uk/~ucgtrbd/code/iv_discovery}}. The parameters of \emph{num$\_$ivs} and \emph{num$\_$boot} are set to 3 (1 for VitD) and 500, respectively. We report the average result of 500 bootstrapping as the final estimated $\hat{\beta}_{wy}$ for IV.tetrad. The implementation of TSLSCIV is based on the functions \emph{glm} and \emph{ivglm} in the $\mathbf{R}$ packages \emph{stats} and \emph{ivtools}, respectively. TSLS is implemented by using the function \emph{ivreg} in the $\mathbf{R}$ package \emph{AER}~\cite{greene2003econometric}. FIVR is implemented by using the function \emph{instrumental}$\_$\emph{forest} in the $\mathbf{R}$ package \emph{grf}~\cite{athey2019generalized}. The implementation of sisVIVE is based on the function \emph{sisVIVE} in the $\mathbf{R}$ package \emph{sisVIVE}. The significance level is set to 0.05 for rFCI used by \emph{AIViP}.

	\paragraph{Evaluation Metrics.} For the synthetic dataset with the true $\beta_{wy}$, we report the estimation bias, $|(\hat{\beta}_{wy} - \beta_{wy})/\beta_{wy}|*100$ (\%). For the real-world datasets, we empirically evaluate the performance of all estimators with the results reported in the corresponding references since the true $\beta_{wy}$ is not available, and we provide the corresponding 95\% confidence interval (C.I.) of $\hat{\beta}_{wy}$ for all estimators.
	
	\begin{table}[t]
		\centering
		\scriptsize
		\begin{tabular}{|c|c|c|c|c|c|c|}
			\hline
			\multicolumn{7}{|c|}{Group I}           \\ \hline
			$n$ &  \emph{AIViP} & TSLS & TSLSCIV & FIVR & sisVIVE & IV.tetrad  \\ \hline
			2k&\textbf{15.0} & 145.2& 342.2 & 79.2 & 27.3 & 32.4  \\ \hline
			3k&\textbf{6.6}& 143.9 & 340.4& 94.6 & 184.3 & 28.0 \\ \hline
			4k&\textbf{27.5} & 143.6& 343.0& 104.4& 53.4 & 30.6\\ \hline
			5k&\textbf{20.9}  & 143.9& 347.7 & 114.5 & 27.7 & 24.1 \\ \hline
			6k&\textbf{3.9} & 142.2& 344.0 & 117.5 & 55.8 &32.1\\ \hline
			8k&\textbf{11.8} & 144.7& 340.6 & 119.9 & 15.8 & 31.9  \\ \hline
			10k&\textbf{21.0}& 141.8 & 342.6 & 130.7  & 320.6& 30.7 \\ \hline
			12k&\textbf{0.2}& 144.2 &340.4  & 132.7& 23.0 & 29.1 \\ \hline
			15k& 29.8 & 145.2& 344.8 & 141.2 & 142.4& \textbf{25.0} \\ \hline
			18k&36.7 & 144.4 & 342.4 & 142.3 & 312.7& \textbf{30.6}\\ \hline
			20k&\textbf{15.2} & 144.6& 341.4 & 144.8 & 187.7&30.9\\ \hline
			\multicolumn{7}{|c|}{Group II}    \\ \hline 
			2k & 63.8& 284.4& 884.7 & 534.4 & 199.4 & \textbf{35.8}\\ \hline
			3k& 54.6& 281.5& 840.3 & 538.9&364.5& \textbf{40.2}\\ \hline
			4k& 47.0& 286.1 & 813.9 & 529.5& 327.8 & \textbf{33.5}\\ \hline
			5k&\textbf{18.2} & 289.7& 838.5 & 571.4 & 396.4& 35.7 \\ \hline
			6k&\textbf{31.5}& 283.1& 837.1 & 581.7& 353.2 & 39.5\\ \hline
			8k& \textbf{26.7} & 285.4& 836.7& 593.0 & 584.6 & 40.1 \\ \hline
			10k&41.5 & 280.5& 807.6 & 572.5& 653.1& \textbf{37.0} \\ \hline
			12k& \textbf{28.6} & 286.3& 818.6 & 588.7& 696.0& 35.3 \\ \hline
			15k& 40.1& 285.4& 824.5 & 604.4 & 652.8& \textbf{35.0} \\ \hline
			18k&\textbf{2.6} & 284.1& 829.8& 612.0 & 823.6&38.8 \\ \hline
			20k&\textbf{16.4} & 291.0& 821.9 & 608.8 & 634.8 & 39.8\\ \hline   
		\end{tabular}
		\caption{Summary of the estimation bias (\%) on both groups of synthetic datasets. The smallest estimation bias on each group is boldfaced. \emph{AIViP} consistently obtains good performance on all datasets.}
		\label{table:synresults}
	\end{table}
	
	\subsection{Simulation Study}
	\label{subsec:Simu}
	We conduct simulation studies to evaluate the performance of \emph{AIViP} when $W$ and $Y$ share a latent confounder $U$. We generate two groups of synthetic datasets with a range of sample sizes: 2k (\ie 2,000), 3k, 4k, 5k, 6k, 8k, 10k, 12k, 15k, 18k, and 20k. The set of observed variables $\mathbf{X}$ is $\{X_1, X_2, \dots, X_{23}, S\}$. We add two and three latent variables for Group I and Group II datasets respectively. The generated synthetic datasets satisfy the three conditions of ancestral IV in Definition~\ref{def:ancestralIV}. The details of the data generating process are provided in Appendix C. To make the results reliable, each reported result is the average of 20 repeated simulations. The estimation biases of all estimators on both groups of synthetic datasets are reported in Table~\ref{table:synresults}.
	
	\paragraph{Results.} From Table~\ref{table:synresults}, we have the following observations: (1) the large estimation biases of TSLS show that the confounding bias caused by the latent confounders between $S$ and $Y$ is not controlled by TSLS at all. (2) TSLSCIV has the largest estimation biases on both groups of synthetic datasets, which shows that conditioning on all variables is inappropriate since the data contains collider bias. (3) The estimation biases of \emph{AIViP} on both groups of datasets show that \emph{AIViP} outperforms FIVR and sisVIVE. This is because both methods fail to detect either colliders or confounding bias in the data. (4) \emph{AIViP} slightly outperforms IV.tetrad in Group I datasets and the two methods have similar performance in Group II datasets. Note that IV.tetrad performs well with synthetic datasets, but not with real-world datasets since its data distribution assumption may not be satisfied in real-world datasets.

	\begin{figure*}[ht]
		\centering
		\includegraphics[scale=0.523]{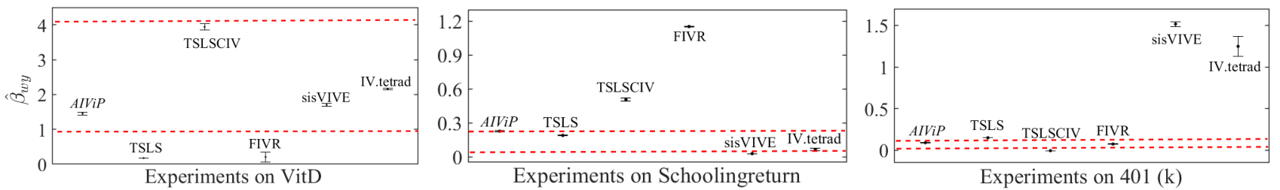}
		\caption{The estimated $\hat{\beta}_{wy}$ on the three real-world datasets. The two dotted lines indicate empirical 95\% confidence interval of the references. Note that the performance of \emph{AIViP} is consistent with the empirical values of the causal effects on all three real-world datasets.}
		\label{fig:threereal-world}
	\end{figure*}
	
	\subsection{Experiments on Real-World Datasets}
	In our experiments, we need to choose some datasets for which the empirical estimates are widely acceptable since there are no ground truths for the real-world datasets. Hence, we evaluate the performance of \emph{AIViP} on three real-world datasets, Vitamin D data (VitD)~\cite{martinussen2019instrumental}, Schoolingreturn~\cite{card1993using} and 401(k) data~\cite{verbeek2008guide}. These datasets are widely utilized in the assessment of IV methods. Each of the three datasets has a nominated conditional IV for estimating the causal effects, but there is not enough knowledge to determine the conditioning sets for the nominated conditional IVs. The details of the three datasets are introduced in Appendix C.
	
	VitD contains 2,571 individuals and 5 variables: age, filaggrin (an instrument), vitd (the treatment variable), time (follow-up time), and death (the outcome variable)~\cite{sjolander2019instrumental}. We take the estimated $\hat{\beta}_{wy} = 2.01$ with 95\% C.I. (0.96, 4.26) from the work~\cite{martinussen2019instrumental} as the reference causal effect.
	
	Schoolingreturn contains 3,010 individuals and 19 variables~\cite{card1993using}. The treatment is the education of employees. The outcome is raw wages in 1976 (in cents per hour). A goal of the study is to investigate the causal effect of education on earnings. Card~\cite{card1993using} uses geographical proximity to a college, \ie \emph{nearcollege} as an instrument variable. We take $\hat{\beta}_{wy} = 13.29$\% with 95\% C.I. (0.0484, 0.2175) from~\cite{verbeek2008guide} as the reference causal effect.
	
	401(k) contains 9,275 individuals from the survey of income and program participation (SIPP) conducted in 1991~\cite{verbeek2008guide,abadie2003semiparametric}. The program participation is about the most popular tax-deferred programs, \ie individual retirement accounts (IRAs) and 401 (k) plans. There are 11 variables about the eligibility for participating in the 401 (k) plan. The treatment is \emph{p401k} (an indicator of participation in 401 (k)), and \emph{pira} (a binary variable, $pira$ = 1 denotes participation in IRA) is the outcome of interest. \emph{e401k} is used as an instrument for \emph{p401k} (an indicator of eligibility for 401 (k)). We take $\hat{\beta}_{wy} = 7.12$\% with 95\% C.I. (0.047, 0.095)~\cite{verbeek2008guide} as the reference causal effect.
	
	\paragraph{Results.} All results on the three datasets are visualized in Fig.~\ref{fig:threereal-world}. From Fig.~\ref{fig:threereal-world}, we have the following observations: (1) \emph{AIViP} obtains results consistent with the reference causal effect values since the estimated causal effects are either in or close to the empirical 95\% C.I. of the reference values on all three datasets. (2) The results of each comparison method are consistent with the reference values in at most two datasets. We note that IV.tetrad performs badly and this may attribute to the fact that its strong assumption on data distribution may not be satisfied. (3) \emph{AIViP} has consistent performance across the three datasets, but all other methods' performances are not consistent across the three datasets and this may attribute to their failure in using the correct conditioning sets to reduce biases. The observations show the advantage of \emph{AIViP} since it identifies the conditioning sets for reducing biases and does not have a strong assumption on data distributions.
	
	\section{Related Work}
	\label{sec:Relatedwork}
	The IV method is a powerful tool in causal inference when the treatment and outcome are confounded by latent variables~\cite{angrist1995two,hernan2006instruments}. It is impossible to test whether a variable is a valid standard IV from observational data alone. Assuming that all variables have discrete values, Pearl proposed \emph{the instrumental inequality} to verify whether a variable is a valid IV~\cite{pearl1995testability}. Kuroki and Cai proposed a criterion to find variables that satisfy the conditions of a standard IV in the linear structural model~\cite{kuroki2005instrumental}. They provided a tighter condition than Pearl's~\cite{pearl1995testability}, and the developed method can be applied to data with continuous or discrete variables. Chu \etal~\cite{chu2001semi} proposed the concept of a semi-instrumental variable for a continuous variable. An IV is a semi-instrument, but the converse does not hold. Under the linearity assumption, Zhang et al.~\cite{zhang2020simultaneous} proposed a symbiotic approach to causal discovery and identification by using a quasi-instrumental set. The four works reviewed above are either theoretical solutions or on a dataset with several variables (less than 5). 
	
	Kang \etal proposed a data-driven IV estimator, sisVIVE~\cite{kang2016instrumental}. sisVIVE requires that a set of candidate IVs and a set of observed variables are known and less than 50\% of the candidate IVs are invalid. Hartford \etal proposed a deep learning based estimator to estimate $\beta_{wy}$ from data~\cite{hartford2021valid}. This method also requires that less than 50\% candidate IVs are invalid. Our work is different from these data-driven methods, as our work is about ancestral IVs and how to find a conditioning set from data. 
	
	The most relevant work to ours is the IV.tetrad method~\cite{silva2017learning}. IV.tetrad aims to find a pair of valid conditional IVs $\{S_i, S_j\}$ from data by using the TETRAD constraint with the strong assumption of linear non-Gaussian causal models. In IV.tetrad, all observed variables in $\mathbf{X}$ excluding $S_i$ and $S_j$ are included in the conditional set $\mathbf{Z}$ that instrumentalizes $S_i$ and $S_j$ simultaneously. This assumption does not always satisfied and this limits the usefulness of IV.tetrad (as shown in our experiments). Different from IV.tetrad, we focus on finding a conditioning set $\mathbf{Z}$ that instrumentalizes a given ancestral IV $S$, to enable the practical use of conditional IVs.

	\section{Conclusion}
	\label{sec:concs}
	One of the major challenges for the real-world application of causal effect estimation is the latent variables in a system, especially when the treatment and outcome share latent confounders. In this work, we study the graphical properties of an ancestral IV using a MAG to estimate causal effect from data with latent variables, including latent confounders. We have proposed the theory for supporting the search for a set of observed variables (a conditioning set) that instrumentalizes a given ancestral IV in a mapped MAG, as well as in a PAG for data-driven discovery of a conditioning set of a given ancestral IV. Based on the theory, we propose an algorithm, \emph{AIViP} to achieve unbiased causal effect estimation from data with latent variables. The extensive experiments on synthetic and real-world datasets demonstrate that \emph{AIViP} is very capable of handling data with latent confounders, even when the data contains collider bias, and \emph{AIViP} outperforms the state-of-the-art estimators.

	\section*{Acknowledgements}We wish to acknowledge the support from the Australian Research Council under DP200101210. JZ's research was supported in part by the RGC of Hong Kong under GRF13602720 and a start-up fund from HKBU.
	
	\bibliographystyle{named}
	\bibliography{ijcai22}

\appendix
\section*{Appendix}

In this Appendix, we provide additional graphical notations and definitions, all proofs of the theorems, and details of synthetic and real-world datasets. 
\section{Background}
%\subsection{Graphical notation and definitions}
\label{subsec:graph}
\textbf{Edges and graphs}. There are three types of end marks for edges in a graph $\mathcal{G}$: arrowhead $(<)$, tail $(-)$, and circle $(\circ)$ (indicating the orientation of the edge is uncertain)~\cite{zhang2008completeness}. An edge has two edge marks and can be directed $\rightarrow$, bi-directed $\leftrightarrow$, non-directed $\cirlinecir$, or partially directed $\circlerightarrow$. A \emph{directed graph} contains only directed edges ($\rightarrow$). A \emph{mixed graph} may contain both directed and bi-directed edges ($\leftrightarrow$)~\cite{zhang2008causal,perkovic2018complete}. A \emph{partial mixed graph} may contain any types of the edges. Noting that we do not consider selection variable (\ie selection bias)~\cite{zhang2008causal}, so non-directed $\cirlinecir$ will not appear in this work.

\textbf{Paths}. In a graph $\mathcal{G}$, a path $\pi$ between $V_{1}$ and $V_{p}$ comprises a sequence of distinct nodes $\langle V_{1}, \dots, V_{p}\rangle$ with every pair of successive nodes being adjacent. A node $V$ lies on the path $\pi$ if $V$ belongs to the sequence $\langle V_{1}, \dots, V_{p}\rangle$. A path $\pi$ is a directed or causal path if all edges along it are directed such as $V_{1} \rightarrow\ldots \rightarrow V_{p}$. In a partial mixed graph, a possibly directed path from $V_i$ to $V_j$ is a path from $V_i$ to $V_j$ that does not contain an arrowhead pointing in the direction to $V_i$. We also refer to this a possibly causal path. A path that does not possibly causal is referred to a non-causal path.

\textbf{Ancestral relationships}. In a directed or mixed graph, $V_i$ is a parent of $V_j$ (and $V_j$ is a child of $V_i$) if $V_i \rightarrow V_j$ appears in the graph.  In a directed path $\pi$, $V_i$ is an ancestor of $V_j$ and $V_j$ is a descendant of $V_i$ if all arrows along $\pi$ point to $V_j$. If there is $V_{i}\leftrightarrow V_{j}$, $V_i$ and $V_j$ are called spouses to each other. If there exists a possibly directed path from $V_i$ to $V_j$, $V_i$ is a possible ancestor of $V_j$, and $V_j$ is a possible descendant of $V_i$.

\textbf{Shields and definite status paths}. A subpath $\langle V_i, V_j, V_k\rangle$ is an unshielded triple if $V_i$ and $V_k$ are not adjacent~\cite{zhang2008causal}. Otherwise, the subpath $\langle V_i, V_j, V_k\rangle$ is a shielded triple. A path is unshielded if all successive triples on the path is unshileded~\cite{perkovic2018complete}. A node $V_j$ is a \emph{definite non-collider} on $\pi$ if there exists at least an edge out of $V_j$ on $\pi$, or both edges have a circle mark at $V_j$ and . A node is of a \emph{definite status} on a path if it is a collider or a definite non-collider on the path. A path $\pi$ is of a \emph{definite status} if every non-endpoint node on $\pi$ is of a definite status~\cite{perkovic2018complete}. 

In graphical causal modelling, the assumptions of Markov property, faithfulness and causal sufficiency are often involved to discuss the relationship between the causal graph and the distribution of the data.

\begin{definition}[Markov property~\cite{pearl2009causality}]
	\label{Markov condition}
	Given a DAG $\mathcal{G}=(\mathbf{V}, \mathbf{E})$ and the joint probability distribution of $\mathbf{V}$ $(P(\mathbf{V}))$, $\mathcal{G}$ satisfies the Markov property if for $\forall V_i \in \mathbf{V}$, $V_i$ is probabilistically independent of all of its non-descendants, given $Pa(V_i)$.
\end{definition}

\begin{definition}[Faithfulness~\cite{spirtes2000causation}]
	\label{Faithfulness}
	A DAG $\mathcal{G}=(\mathbf{V}, \mathbf{E})$ is faithful to a joint distribution $P(\mathbf{V})$ over the set of variables $\mathbf{V}$ if and only if every independence present in $P(\mathbf{V})$ is entailed by $\mathcal{G}$ and satisfies the Markov property. A joint distribution $P(\mathbf{V})$ over the set of variables $\mathbf{V}$ is faithful to the DAG $\mathcal{G}$ if and only if the DAG $\mathcal{G}$ is faithful to the joint distribution $P(\mathbf{V})$.
\end{definition}

\begin{definition}[Causal sufficiency~\cite{spirtes2000causation}]
	A given dataset satisfies causal sufficiency if in the dataset for every pair of observed variables, all their common causes are observed.
\end{definition}

In a DAG, d-separation is a graphical criterion that enables the identification of conditional independence between variables entailed in the DAG when the Markov property, faithfulness and causal sufficiency are satisfied~\cite{pearl2009causality,spirtes2000causation}.

\begin{definition}[d-separation~\cite{pearl2009causality}]
	\label{d-separation}
	A path $\pi$ in a DAG $\mathcal{G}=(\mathbf{V}, \mathbf{E})$ is said to be d-separated (or blocked) by a set of nodes $\mathbf{Z}$ if and only if
	(i) $\pi$ contains a chain $V_i \rightarrow V_k \rightarrow V_j$ or a fork $V_i \leftarrow V_k \rightarrow V_j$ such that the middle node $V_k$ is in $\mathbf{Z}$, or
	(ii) $\pi$ contains a collider $V_k$ such that $V_k$ is not in $\mathbf{Z}$ and no descendant of $V_k$ is in $\mathbf{Z}$.
	A set $\mathbf{Z}$ is said to d-separate $V_i$ from $V_j$ ($ V_i \indep_d V_j\mid\mathbf{Z}$) if and only if $\mathbf{Z}$ blocks every path between $V_i$ to $V_j$. otherwise they are said to be d-connected by $\mathbf{Z}$, denoted as $V_i\nindep_d V_j\mid\mathbf{Z}$.
\end{definition}

Ancestral graphs as defined below are often used to represent the mechanisms of data generating process that may involve latent variables~\cite{richardson2002ancestral}.

\begin{definition}[Ancestral graph]
	\label{Ancestral graph}
	An ancestral graph is a mixed graph that does not contain directed cycles or almost directed cycles.
\end{definition}

\begin{figure}[t]
	\centering
	\includegraphics[width=6.8cm]{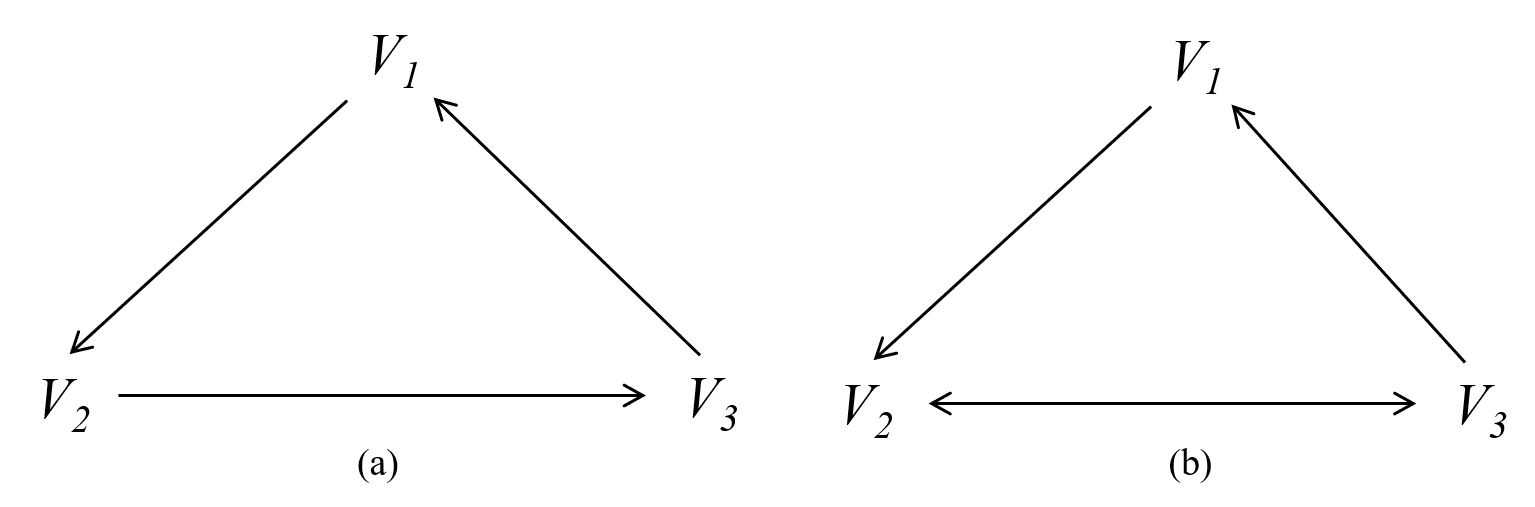}
	\caption{An example of a direct cycle and almost cycle.}
	\label{fig:001}
\end{figure}

The direct cycle and almost cycle are two important concepts in an ancestral graph. Here, we provide an example in Fig.~\ref{fig:001} to show the direct cycle and almost cycle.

The criterion of m-separation is a natural extension of the d-separation criterion to ancestral graphs.
\begin{definition}[m-separation~\cite{spirtes2000causation}]
	\label{m-separation}
	In an ancestral graph $\mathcal{M}=(\mathbf{V}, \mathbf{E})$, a path $\pi$ between $V_{i}$ and $V_{j}$ is said to be m-separated by a set of nodes $\mathbf{Z}\subseteq \mathbf{V}\setminus\{V_i, V_j\}$ (possibly $\emptyset$) if $\pi$ contains a subpath $\langle V_l, V_k, V_s\rangle$ such that the middle node $V_k$ is a non-collider on $\pi$ and $V_k \in \mathbf{Z}$; or $\pi$ contains $V_l\astrightarrow V_k\astleftarrow V_s$ such that $V_k \notin \mathbf{Z}$ and no descendant of $V_k$ is in $\mathbf{Z}$. Two nodes $V_{i}$ and $V_{j}$ are said to be m-separated by $\mathbf{Z}$ in $\mathcal{M}$, denoted as $V_i\indep_m V_j\mid\mathbf{Z}$ if every path between $V_{i}$ and $V_{j}$ are m-separated by $\mathbf{Z}$; otherwise they are said to be m-connected by $\mathbf{Z}$, denoted as $V_i\nindep_m V_j\mid\mathbf{Z}$.
\end{definition}

\begin{figure}[t]
	\centering
	\includegraphics[width=7.55cm]{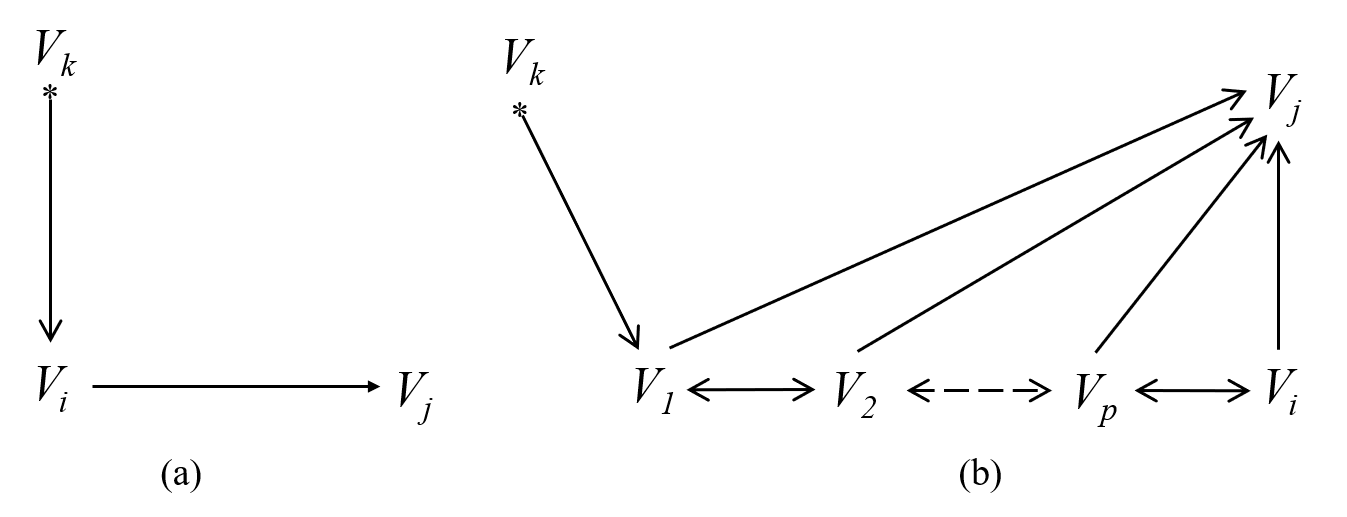}
	\caption{Two possible configurations of the visible edge between $V_i \rightarrow V_j$. Note that $V_k$ and $V_j$ are non-adjacent.}
	\label{fig:visibleedge}
\end{figure}

The visible edge (Definition 3 in the main text) is a critical concept in a MAG/PAG, so two possible configurations of the visible edge $V_i$ to $V_j$ are provided as shown in Fig.~\ref{fig:visibleedge}.

A DAG over observed and unobserved variables can be converted to a MAG with observed variables. From a DAG over $\mathbf{X}\cup\mathbf{U}$ where $\mathbf{X}$ is a set of observed variables and $\mathbf{U}$ is a set of unobserved variables, following the construction rule specified in~\cite{zhang2008completeness}, one can construct a MAG with nodes $\mathbf{X}$ such that all the conditional independence relationships among the observed variables entailed by the DAG are entailed by the MAG and vice versa, and the ancestral relationships in the DAG are maintained in the MAG.

Inducing path is necessary to convert a DAG to a MAG.

\begin{definition}[Inducing path~\cite{richardson2002ancestral,zhang2008completeness}]
	In an ancestral graph $\mathcal{G}$, let $X$ and $Y$ be two nodes, and $\mathbf{U}$ be a set of nodes not containing $X, Y$. A path $\pi$ from $X$ to $Y$ is called an \textbf{inducing path} \wrt $\mathbf{U}$ if every non-endpoint node on $\pi$ is either in $\mathbf{U}$ or a collider, and every collider on $\pi$ is an ancestor of either $X$ or $Y$. When $\mathbf{U}=\emptyset$, $\pi$ is called a \textbf{primitive inducing path }from $X$ to $Y$.
\end{definition}

The construction rules of a MAG $\mathcal{M}$ over $\mathbf{X}$ from a given DAG $\mathcal{G}$ over $\mathbf{X}\cup\mathbf{U}$~\cite{zhang2008completeness} are provided as follow.  
\begin{description}
	\item[Input:]  a DAG $\mathcal{G}$ over $\mathbf{X}\cup\mathbf{U}$
	\item[Output:]  a MAG $\mathcal{M}$ over $\mathbf{X}$ \\
	(1) For each pair of variables $X, Y\in \mathbf{X}$, $X$ and $Y$ are adjacent in $\mathcal{M}$ \ifff there is an inducing path from $X$ to $Y$ \wrt $\mathbf{U}$ in $\mathcal{G}$. \\
	(2) For each pair of adjacent nodes $X$ and $Y$ in $\mathcal{M}$, orient the edge between them as follows.
	\begin{description}
		\item[(a)] $X\rightarrow Y$ if $X\in An(Y)$ and $Y\notin An(X)$;
		\item[(b)] $X\leftarrow Y$ if $X\notin An(Y)$ and $Y\in An(X)$;
		\item[(c)] $X\leftrightarrow Y$ if $X\notin An(Y)$ and $Y\notin An(X)$.
	\end{description}
\end{description}

If two MAGs represent the same set of m-separations, they are called \emph{Markov equivalent}, and formally, we have the following definition.

\begin{definition}[Markov equivalent MAGs~\cite{zhang2008completeness}]
	\label{markovequ}
	Two MAGs $\mathcal{M}_1$ and $\mathcal{M}_2$ with the same nodes are said to be \emph{Markov equivalent}, denoted $\mathcal{M}_1 \sim \mathcal{M}_2$, if for all triple nodes $X$, $Y$, $Z$, $X$ and $Y$ are m-separated by $Z$ in $\mathcal{M}_1$ if and only if $X$ and $Y$ are m-separated by $Z$ in $\mathcal{M}_2$.
\end{definition}

The set of all MAGs that encode the same set of m-separations form a \emph{Markov equivalence class}~\cite{spirtes2000causation}. 
A set of Markov equivalent MAGs can be represented by a PAG and defined as below.
\begin{definition}[PAG]
	\label{def:pag}
	Let $[\mathcal{M}]$ be the Markov equivalence class of a MAG $\mathcal{M}$. The PAG $\mathcal{P}$ for $[\mathcal{M}]$ is a partially mixed graph if (i). $\mathcal{P}$ has the same adjacent relations among nodes as $\mathcal{M}$ does; (ii). For an edge, its mark of arrowhead or mark of the tail is in $\mathcal{P}$ if and only if the same mark of arrowhead or the same mark of the tail is shared by all MAGs in $[\mathcal{M}]$.
\end{definition}

\section{Finding a conditioning set for an ancestral IV in data}
\label{sec:theo_appendix}
\subsection{Representing an ancestral IV in MAG}
\label{Subsec:relaxedIV_appendix}

\textbf{Lemma 1.} \emph{Given a DAG $\mathcal{G} = (\mathbf{X}\cup\mathbf{U}\cup\{W, Y\}, \mathbf{E}')$ with the edges $W\rightarrow Y$ and $W\leftarrow U\rightarrow Y$ in $\mathbf{E}'$, and $U\in \mathbf{U}$, and let a MAG $\mathcal{M} = (\mathbf{X}\cup\{W, Y\}, \mathbf{E})$ be mapped from $\mathcal{G}$ based on the construction rules~\cite{zhang2008completeness}. Suppose that there exists an ancestral IV $S$ conditioning on a set  $\mathbf{Z}\subseteq\mathbf{X}\setminus\{S\}$ in $\mathcal{G}$. In the mapped MAG $\mathcal{M}$, the edge $W\rightarrow Y$ is invisible and there is an edge $S\rightarrow Y$ or $S\leftrightarrow Y$.}
\begin{proof}
In the DAG $\mathcal{G}$, $W\rightarrow Y$ and there is an inducing path $W\leftarrow U\rightarrow Y$ relative to $U$. Thus, in the mapped MAG $\mathcal{M}$, there is a directed edge $W\rightarrow Y$ that is an invisible edge (Lemma 9 in~\cite{zhang2008causal}).
	
	$S$ is an ancestral IV in $\mathcal{G}$, so, there exists $\mathbf{Z}\subseteq\mathbf{X}\setminus\{S\}$ such that $S\nindep_{d} W\mid\mathbf{Z}$ in $\mathcal{G}$ and $S\indep_{d} Y\mid\mathbf{Z}$ in $\mathcal{G}_{\underline{W}}$ according to the ancestral IV in DAG (Definition~7 in the main text). Moreover, $S\nindep_{d} Y\mid \mathbf{Z}$ holds for $\forall \mathbf{Z}\subseteq \mathbf{X}\setminus\{S\}$ in $\mathcal{G}_{\underline{W}}$ due to the latent variable $U$ between $W$ and $Y$ in $\mathcal{G}$. That is, $S$ and $Y$ are adjacent in the mapped MAG $\mathcal{M}$. Therefore, if $S\in An(Y)$ in $\mathcal{G}$, then the edge between $S$ and $Y$ is oriented as $S\rightarrow Y$ in $\mathcal{M}$. Otherwise the edge is oriented as $S\leftrightarrow Y$ in $\mathcal{M}$ according to the construction rules.
\end{proof}

\subsection{The property of an ancestral IV in MAG}
\label{subsec:property_ANIV_appendix}
\textbf{Lemma 2.} [The property of an ancestral IV in MAG].
	\emph{Given a DAG $\mathcal{G} = (\mathbf{X}\cup\mathbf{U}\cup\{W, Y\}, \mathbf{E}')$ with the edges $W\rightarrow Y$ and $W\leftarrow U\rightarrow Y$ in $\mathbf{E}'$, and $U\in\mathbf{U}$, and let MAG $\mathcal{M} = (\mathbf{X}\cup\{W, Y\}, \mathbf{E})$ be mapped from $\mathcal{G}$. Suppose that there exists an ancestral IV $S$ conditioning on a set $\mathbf{Z}\subseteq\mathbf{X}\setminus\{S\}$ in $\mathcal{G}$. In the mapped MAG $\mathcal{M}$, if a set $\mathbf{Z}\subseteq\mathbf{X}\setminus\{S\}$ satisfies (i) $S$ and $W$ are not m-separated given $\mathbf{Z}$ in $\mathcal{M}$, and (ii) $S$ and $Y$ are m-separated by $\mathbf{Z}$ in $\mathcal{M}_{\underline{W}\widetilde{S}}$, then $\mathbf{Z}$ instrumentalizes $S$ in the DAG $\mathcal{G}$.}
\begin{proof}
 There exists an edge between $S$ and $Y$ in the mapped MAG $\mathcal{M}$ according to Lemma~\ref{lemma:001}. The edge between $S$ and $Y$ is added in $\mathcal{M}$ to represent the spurious association caused by the latent confounder $U$, so removing it from the mapped MAG will not change the causal relationships between $S$ and $Y$. The manipulated MAG by removing the edge between $S$ and $Y$ is denoted as $\mathcal{M}_{\widetilde{S}}$.  Furthermore, the manipulated MAG $\mathcal{M}_{\underline{W}}$ is constructed by replacing the edge $W\rightarrow Y$ with $W\leftrightarrow Y$ since $W\rightarrow Y$ is an invisible edge (according to manipulations of MAGs in Definition 11 of the literature~\cite{zhang2008causal}).

  Because (i) $S$ and $W$ are not m-separated given $\mathbf{Z}$ in $\mathcal{M}$, then there is a d-connection path between $S$ and $W$ in the DAG $\mathcal{G}$, \ie (a) $S\indep_{d} W\mid \mathbf{Z}$ in $\mathcal{G}$. Because (ii) $S$ and $Y$ are m-separated by $\mathbf{Z}$ in $\mathcal{M}_{\underline{W}\widetilde{S}}$, \ie all paths from $S$ to $Y$ are blocked by $\mathbf{Z}$ in $\mathcal{M}_{\underline{W}\widetilde{S}}$, so $S$ and $Y$ are d-separated by $\mathbf{Z}$ in $\mathcal{G}_{\underline{W}}$, \ie (b) $S\indep_{d} Y\mid \mathbf{Z}$ in $\mathcal{G}_{\underline{W}}$.  Under the pretreatment assumption, there is not a descendant node of $Y$, \ie (c) $\forall Z\in \mathbf{Z}$, $Z\notin De(Y)$. Therefore, $\mathbf{Z}$ instrumentalizes $S$ in the DAG $\mathcal{G}$ because of (a), (b) and (c).
\end{proof}

\subsection{Determining a conditioning set in a MAG}
\label{Subsec:relaxedIV_appendix}

\textbf{Corollary 1.} \emph{Given a DAG $\mathcal{G} = (\mathbf{X}\cup\mathbf{U}\cup\{W, Y\}, \mathbf{E}')$ with the edges $W\rightarrow Y$ and $W\leftarrow U\rightarrow Y$ in $\mathbf{E}'$, and $U\in\mathbf{U}$, and let MAG $\mathcal{M} = (\mathbf{X}\cup\{W, Y\}, \mathbf{E})$ be mapped from $\mathcal{G}$. For a given ancestral IV $S$, $D$-$SEP(S, Y)$ in the mapped MAG $\mathcal{M}$ is a set that instrumentalizes $S$ in the DAG $\mathcal{G}$.}
\begin{proof}
	$D$-$SEP(S, Y)$ contains $An(S)$ and $An(Y)$ according to Definition~\ref{def:DSEP}, so $D$-$SEP(S, Y)$ satisfies the clause (iii) of Definition~\ref{def:ancestralIV}. In the mapped MAG $\mathcal{M}$, the edge $W\rightarrow Y$ is invisible, so $S$ and $Y$ are m-connection given $Pa(Y)$ (possibly empty). $S$ and $W$ are not m-separated given $D$-$SEP(S, Y)$ in $\mathcal{M}$ since $Pa(Y)\subseteq D$-$SEP(S, Y)$. Hence, $D$-$SEP(S, Y)$ satisfies the condition (i) of Lemma 2. Moreover, $S$ and $Y$ are non-adjacent in $\mathcal{M}_{\widetilde{S}}$, so $S$ and $Y$ are m-separated by $D$-$SEP(S, Y)$ in $\mathcal{M}_{\underline{W}\widetilde{S}}$ based on Lemma~3. Hence, $D$-$SEP(S, Y)$ satisfies both conditions of Lemma 2. Therefore, $D$-$SEP(S, Y)$ in the mapped $\mathcal{M}$ is a set that instrumentalizes $S$ in the DAG $\mathcal{G}$.
\end{proof}

\subsection{Determining a conditioning set $\mathbf{Z}$ in PAG}
\label{subsec:conditioningset_PAG_appendix}
	\textbf{Theorem 1.} Given a DAG $\mathcal{G} = (\mathbf{X}\cup\mathbf{U}\cup\{W, Y\}, \mathbf{E}')$ with the edges $W\rightarrow Y$ and $W\leftarrow U\rightarrow Y$ in $\mathbf{E}'$, and $U\in\mathbf{U}$, and let MAG $\mathcal{M} = (\mathbf{X}\cup\{W, Y\}, \mathbf{E})$ be mapped from $\mathcal{G}$. From data, the mapped MAG $\mathcal{M}$ is represented by a PAG $\mathcal{P}=(\mathbf{X}\cup\{W, Y\}, \mathbf{E}'')$. For a given ancestral IV $S$ which is a cause or spouse of $W$, the set $PossAn(S\cup Y)\setminus\{W, S\}$ in the learned $\mathcal{P}$ is a set that instumentalizes $S$ in the DAG $\mathcal{G}$.
\begin{proof}
In the mapped MAG $\mathcal{M}$, there exists an edge between $S$ and $Y$ based on Lemma~\ref{lemma:001}. So there is still an edge between $S$ and $Y$ in the learned PAG $\mathcal{P}$ due to the mapped MAG $\mathcal{M}$ is represented in $\mathcal{P}$. Thus the edge between $S$ and $Y$ is due to the spurious association caused by the latent confounder $U$ and can be removed without changing any causal information between $S$ and $Y$ in $\mathcal{P}$, and denoted as $\mathcal{P}_{\widetilde{S}}$. The manipulated PAG $\mathcal{P}_{\underline{W}}$ is constructed by replacing the edge $W\circlerightarrow Y$ with $W\leftrightarrow Y$ in $\mathcal{P}$ since the edge $W\circlerightarrow Y$ is not definite visible according to manipulations of PAGs in Definition 15 of~\cite{zhang2008causal}. So, the manipulated PAG $\mathcal{P}_{\underline{W}\widetilde{S}}$ is obtained by replacing $W\circlerightarrow Y$ with $W\leftrightarrow Y$ and removing the edge between $S$ and $Y$. Thus, $S$ and $Y$ are non-adjacent in the manipulated PAG $\mathcal{P}_{\underline{W}\widetilde{S}}$.
	
In our problem setting, the ancestral IV $S$ is a cause or spouse of $W$, \ie $S$ and $W$ are m-connection given any set $\mathbf{Z}$ in the mapped MAG $\mathcal{M}$, then the set $PossAn(S\cup Y)\setminus\{W, S\}$ discovered in $\mathcal{P}$ satisfies $S\nindep_m W\mid PossAn(S\cup Y)\setminus\{W, S\}$ in $\mathcal{M}$. Hence, (a) $PossAn(S\cup Y)\setminus\{W, S\}$ satisfies the condition (i) of Lemma 2. 
		
Next we are going to proof that $S$ and $Y$ are m-separated by $PossAn(S\cup Y)\setminus\{W, S\}$ in $\mathcal{P}_{\underline{W}\widetilde{S}}$, \ie $S\indep_m Y\mid PossAn(S\cup Y)\setminus\{W, S\}$,  using contradiction. Suppose that $S\nindep_m Y\mid PossAn(S\cup Y)\setminus\{W, S\}$ in $\mathcal{P}_{\underline{W}\widetilde{S}}$. There will be a m-connection path between $S$ and $Y$ in $\mathcal{P}_{\underline{W}\widetilde{S}}$. The mapped MAG $\mathcal{M}$ is represented in the PAG  $\mathcal{P}$, so $S$ and $Y$ are m-connection given $PossAn(S\cup Y)\setminus\{W, S\}$ in $\mathcal{M}_{\underline{W}\widetilde{S}}$ due to the Markov equivalent. That means, $S$ and $Y$ are d-connection conditioning on  $PossAn(S\cup Y)\setminus\{W, S\}$ in the DAG $\mathcal{G}_{\underline{W}}$ \ie there is not a set $\mathbf{Z}\subseteq\mathbf{X}\setminus\{S\}$ for the given ancestral IV $S$. This contradicts with Definition~\ref{def:ancestralIV} ancestral IV, \ie $S$ is not a given ancestral IV. Hence, $S$ and $Y$ are m-separated by $PossAn(S\cup Y)\setminus\{W, S\}$ in $\mathcal{M}_{\underline{W}\widetilde{S}}$, \ie (b) $PossAn(S\cup Y)\setminus\{W, S\}$ satisfies the condition (ii) of Lemma 2. Therefore, $PossAn(S\cup Y)\setminus\{W, S\}$ is a set in the PAG $\mathcal{P}$ that instrumentalizes $S$ in the DAG $\mathcal{G}$ because of (a) and (b).
\end{proof}

\section{Experiments}
\subsection{Synthetic datasets}
\begin{figure}[t]
	\centering
	\includegraphics[width=6.9cm]{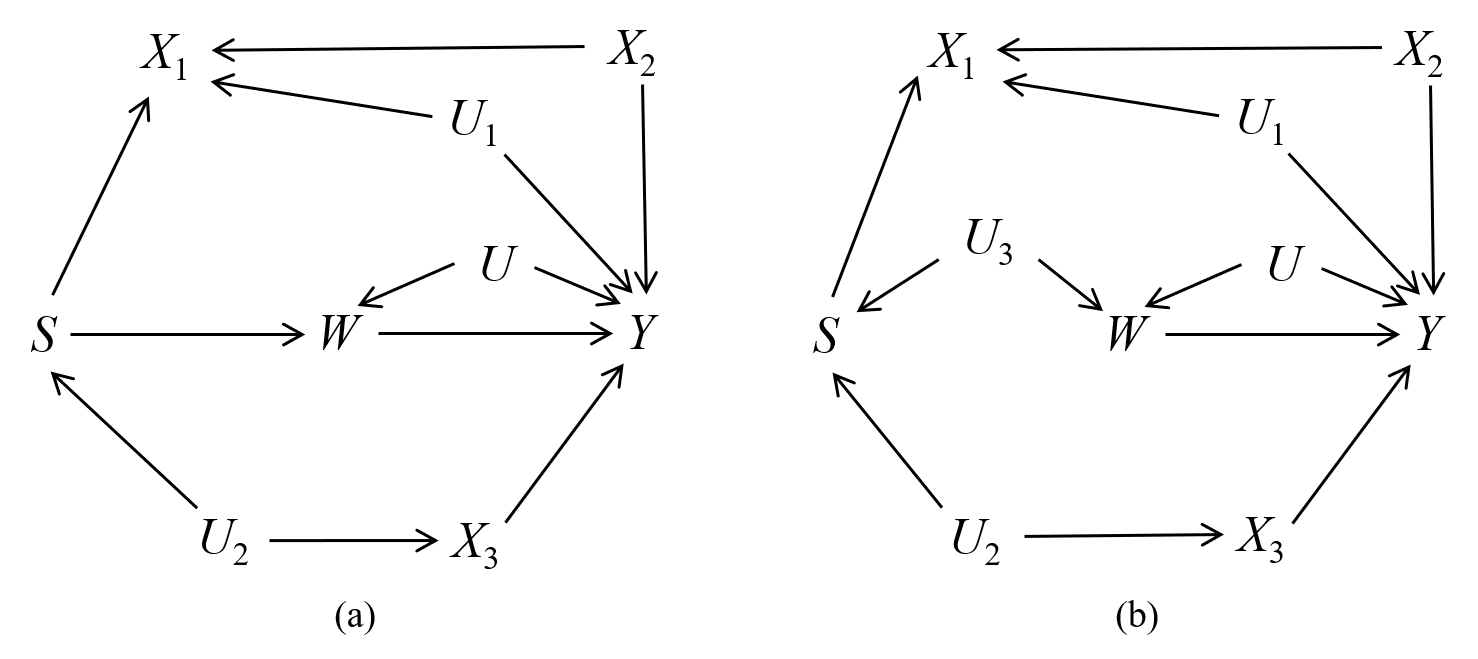}
	\caption{The two true DAGs over $\mathbf{X}\cup\mathbf{U}\cup\{W, Y\}$ are used to generate the synthetic datasets. In DAG (a), $S$ is a cause of $W$, and in DAG (b) $S$ is a spouse of $W$.}
	\label{fig:TheTwoTrueDAGs}
\end{figure}

We utilize two true DAGs over $\mathbf{X}\cup\mathbf{U}\cup\{W, Y\}$ to generate two groups of the synthetic datasets. The two true DAGs are shown in Fig.~\ref{fig:TheTwoTrueDAGs}. The only difference between the two true DAGs is the causal relationship between the ancestral IV $S$ and the treatment $W$. In DAG (a) of Fig.~\ref{fig:TheTwoTrueDAGs}, $S$ is a cause of $W$, while in DAG (b) of Fig.~\ref{fig:TheTwoTrueDAGs}, $S$ is a spouse of $W$ (\ie there is no causal relationship between $S$ and $W$).

In addition to the variables in the two true DAGs, 20 additional observed variables are generated as noise variables that are related to each other but not to the nodes in the two DAGs. Hence, the set of observed covariates is $\mathbf{X}=\{X_1, X_2, \dots, X_{23}, S\}$. The set of unobserved variables is $\mathbf{U}=\{U, U_1, U_2\}$ for Group I and $\mathbf{U}=\{U, U_1, U_2, U_3\}$ for Group II, respectively. $S$ and $\mathbf{Z}=\{X_3\}$ satisfy the three conditions of ancestral IV in the two true DAG over $\mathbf{X}\cup\mathbf{U}\cup\{W, Y\}$. It is worth noting that $X_1$ is a collider and collider bias will be introduced if $X_1$ is incorrectly included in $\mathbf{Z}$.

The Group I of synesthetic datasets are generated based on the DAG (a) in Fig.~\ref{fig:TheTwoTrueDAGs}, and the specifications are as following: $U\sim Bernoulli(0.5)$, $U_1, U_2 \sim N(0, 1)$, $\epsilon_{S}, \epsilon_{X_2}, \epsilon_{X_3} \sim N(0, 0.5)$, $S = N(0, 1)+0.8*U_2 + \epsilon_S$, $X_2 \sim N(0, 1)$, $X_1 = 0.3 + S + X_2 + U_1 + \epsilon_{X_2}$, $X_3 = N(0, 1) +0.8*U_2 + \epsilon_{X_3}$, and the rest of covariates, \ie $X_{4}, X_{5}, \dots, X_{23}$ are generated by multivariate normal distribution. Note that $N(,)$ denotes the normal distribution. The treatment $W$ is generated from $n$ ($n$ denotes the sample size) Bernoulli trials by using the assignment probability $P(W=1\mid U, S) = [1+exp\{1-2*U-2*S\}]$. The potential outcome is generated from $Y_{W} = 2 + 2*W + 2*U + 2*X_1 + 2*U_1 +2*X_3 +\epsilon_{w}$ where $\epsilon_{w}\sim N(0,1)$.

The Group II of synesthetic datasets are generated based on the DAG (b) in Fig.~\ref{fig:TheTwoTrueDAGs}, and the specifications are mostly the same as those for generating Group I. The differences are, $U_3 \sim N(0, 1)$, $S = N(0, 1)+0.8*U_2 + 0.8*U_3 \epsilon_S$, and the treatment $W$ is generated based on $n$ Bernoulli trials by $P(W=1\mid U, U_3) = [1+exp\{1-2*U-2*U_3\}]$. 

All data generation and experiments are conducted with $\mathbf{R}$ programming language. All experiments are repeated 20 times, with a range of sample sizes, \ie 2k (stands for 2,000), 3k, 4k, 5k, 6k, 8k, 10k, 12k, 15k, 18k, and 20k.

\subsection{Real-world datasets}
\textbf{Vitamin D data}. VitD is a cohort study of vitamin D status on mortality reported in~\cite{martinussen2019instrumental}. The data contains 2,571 individuals and 5 variables: age, filaggrin (a binary variable indicating filaggrin mutations), vitd (a continuous variable measured as serum 25-OH-D (nmol/L)), time (follow-up time), and death (binary outcome indicating whether an individual died during follow-up)~\cite{sjolander2019instrumental}. The measured value of vitamin D less than 30 nmol/L implies vitamin D deficiency. The indicator of filaggrin is used as an instrument~\cite{martinussen2019instrumental}. We take the estimated $\hat{\sigma}_{wy} = 2.01$ with 95\% C.I. (0.96, 4.26) from the work~\cite{martinussen2019instrumental} as the reference causal effect.

\textbf{Schoolreturning}. The data is from the national longitudinal survey of youth (NLSY), a well-known
dataset of US young employees, aged range from 24 to 34~\cite{card1993using}. The treatment is the education of employees, and the outcome is raw wages in 1976 (in cents per hour). The data contains 3,010 individuals and 19 covariates. The covariates include experience (Years of labour market experience), ethnicity (Factor indicating ethnicity), resident information of an individual, age, nearcollege (whether an individual grew up near a 4-year college?), marital status, Father's educational attainment,
Mother's educational attainment, and so on. A goal of the studies on this dataset is to investigate the causal effect of education on earnings. Card~\cite{card1993using} used geographical proximity to a college, \ie the covariate \emph{nearcollege} as an instrument variable. We take $\hat{\sigma}_{wy} = 13.29$\% with 95\% C.I. (0.0484, 0.2175) from~\cite{verbeek2008guide} as the reference causal effect.

\textbf{401(k) data}. This dataset is a cross-sectional data from the Wooldridge data sets\footnote{\url{http://www.stata.com/texts/eacsap/}}~\cite{wooldridge2010econometric}. The program participation is about the most popular tax-deferred programs, \ie individual retirement accounts (IRAs) and 401 (k) plans. The data contains 9,275 individuals from the survey of income and program participation (SIPP) conducted in  1991~\cite{abadie2003semiparametric}. There are 11 variables about the eligibility for participating in 401 (k) plans, \wrt income and demographic information, including \emph{pira} (a binary variable, \emph{pira} = 1 denotes participation in IRA), \emph{nettfa} (net family financial assets in \$1,000) \emph{p401k} (an indicator of participation in 401(k)), \emph{e401k} (an indicator of eligibility for 401(k)), \emph{inc} (income), \emph{incsq} (income square), \emph{marr} (marital status), \emph{gender}, \emph{age}, \emph{agesq} (age square) and \emph{fsize} (family size). The treatment $W$ is \emph{p401k} and \emph{pira} is the outcome of interest. \emph{e401k} is used as an instrument for $W$ \emph{p401k}~\cite{abadie2003semiparametric}. We take $\hat{\sigma}_{wy} = 7.12$\% with 95 \% C.I. (0.047, 0.095) from~\cite{abadie2003semiparametric} as the reference causal effect.
\subsection{The running times of all methods on three real-world datasets}
$AIViP$ has a similar time complexity as TSLS, TSLSCIV and sisVIVE. We summarized the running times of all IV estimators on three real-world datasets in Table~2. From Table~2, we have that $AIViP$, TSLS, TSLSCIV and sisVIVE take around 1.5 seconds on a dataset in our experiment. FIVR and IV.tetrad take a longer time since they build the random forest and bootstrap a dataset respectively. For example, FIVR takes 5-40 seconds on a dataset when it builds 2000 trees and IV.tetrad takes around 5 seconds on a dataset when it bootstraps a dataset 500 times.

\begin{table}[t]
	\label{runningtimes}
	\begin{tabular}{|c|c|c|c|}
		\hline
		Methods   & VitD & SchoolingReturns & 401 k \\ \hline
		$AIViP$   & 0.08 & 1.25             & 0.39  \\ \hline
		TSLS      & 0.02 & 0.03             & 0.01  \\ \hline
		TSLSCIV   & 0.03 & 0.05             & 0.08  \\ \hline
		FVIR      & 5.58 & 17.48            & 49.2  \\ \hline
		sisVIVE   & 0.07 & 0.16             & 0.55  \\ \hline
		IV.tetrad & 1.14 & 5.14             & 4.98  \\ \hline
	\end{tabular}
\caption{The running times of all IV estimators on three real-world datasets (seconds).}
\end{table}

\end{document}